\pdfoutput=1
\documentclass[preprint,12pt]{elsarticle}

\usepackage{prletters}
\usepackage{framed,multirow}

\usepackage{amssymb}
\usepackage{latexsym}

\usepackage{url}
\usepackage{xcolor}
\definecolor{newcolor}{rgb}{.8,.349,.1}

\usepackage{algorithmic}
\usepackage{booktabs}
\usepackage{multirow}
\usepackage{float}

\usepackage{makecell}
\usepackage{adjustbox}
\usepackage{float}
\floatstyle{plaintop}
\restylefloat{table}
\usepackage{placeins}
\usepackage{subcaption}
\usepackage{relsize}

\usepackage{amsmath,amssymb,amsfonts}
\usepackage{amssymb}

\newcommand{\spnet}{ShapedNet}

\newcommand{\bcdb}{BCDB23}

\renewcommand{\vec}[1]{\mathbf{#1}}

\usepackage[tableposition=top]{caption}

\journal{Pattern Recognition Letters}

\begin{document}

\thispagestyle{empty}

\begin{table*}[!th]

\section*{Graphical Abstract (Optional)}
To create your abstract, please type over the instructions in the
template box below.  Fonts or abstract dimensions should not be changed
or altered. 

\vskip1pc
\fbox{
\begin{tabular}{p{.4\textwidth}p{.5\textwidth}}
\bf ShapedNet: Advancing Body Composition Assessment with Deep Regression  \\
Navar Medeiros M. Nascimento,
Pedro Cavalcante de Sousa Junior,
Pedro Yuri Rodrigues Nunes,
Suane Pires Pinheiro da Silva,
Luiz Lannes Loureiro,
Victor Zaban Bittencourt,
Valden Luis Matos Capistrano Junior,
Pedro Pedrosa Rebouças Filho \\[1pc]
\includegraphics[width=.6\textwidth]{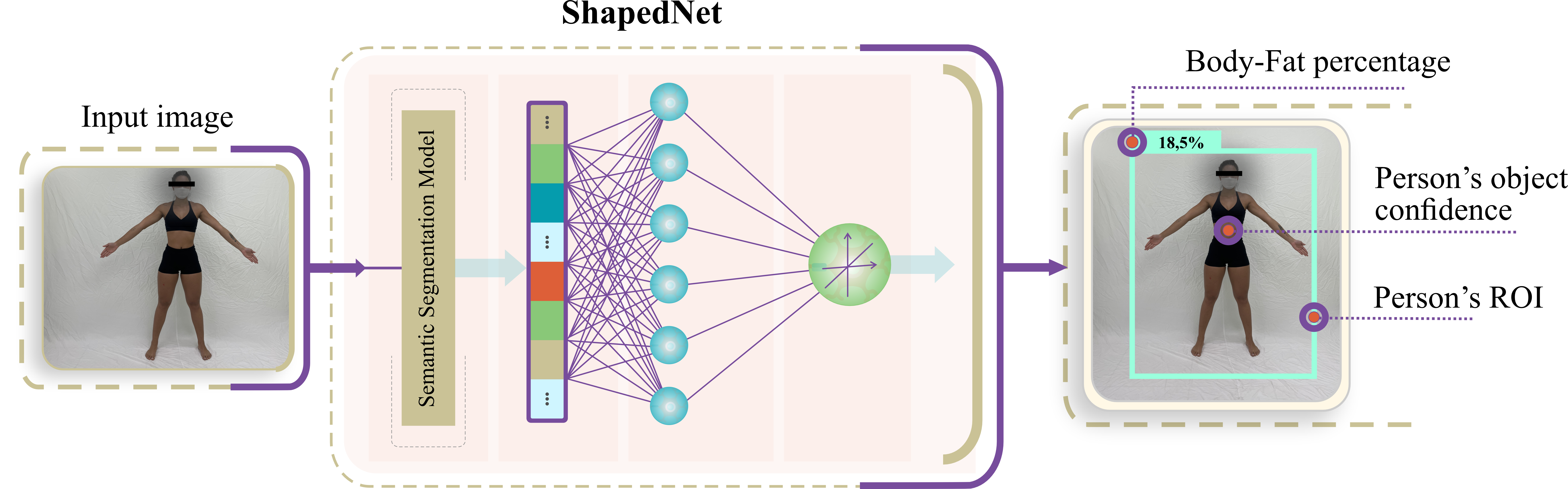}
& 
The goal of our network is not limited to predict human's body fat percentage, but also localize and identify the person's body. 
The whole process is summarized in the figure presented here.
The core ideia is that the feature representations used to identify objects in inner layers empowers of meaningful representation to estimate body fat percentage as well. 
An input image containing a person is given to the model and after a series of operations, such as convolutions, the network provides the following ouputs: (1) the body fat percentage; (2) the person's ROI information, \textit{i.e.} $x, y, w, h$; (3) person's object confidence and (4) the object class, in this case is set to 1, since the network is only trained to identify humans.

\end{tabular}
}

\end{table*}

\clearpage
\thispagestyle{empty}
\vspace*{-1pc}

\begin{table*}[!t]

\section*{Research Highlights (Required)}


\vskip1pc

\fboxsep=6pt
\fbox{
\begin{minipage}{.95\textwidth}
\begin{itemize}

 \item Body fat Estimation

 \item Gold Standard Validation

 \item Single Photo Deep Regression

\end{itemize}
\vskip1pc
\end{minipage}
}

\end{table*}

\clearpage

\setcounter{page}{1}

\begin{frontmatter}

\title{New Advances in Body Composition Assessment with ShapedNet: A Single Image Deep Regression Approach}

\author[1,2]{Navar \surname{Medeiros M. Nascimento}\corref{cor1}} 
\cortext[cor1]{Corresponding author: }
\ead{navarmn@ifce.edu.br}

\author[1]{Pedro \surname{Cavalcante de Sousa Junior}}
\author[1]{Pedro Yuri \surname{Rodrigues Nunes}}
\author[2]{Suane \surname{Pires Pinheiro da Silva}}
\author[3]{Luiz \surname{Lannes Loureiro}}
\author[3]{Victor Zaban \surname{Bittencourt}}
\author[3]{Valden \surname{Luis Matos Capistrano Junior}}
\author[1,2]{Pedro Pedrosa \surname{Rebouças Filho}}


\affiliation[1]{organization={a
Laboratory for Processing Image, Signals and Computer Science (LAPISCO), Federal Institute of Science and Technology of Ceará (IFCE)},
                addressline={Av. Treze de Maio, 2081}, 
                city={Fortaleza}, 
                postcode={60040-531}, 
                state={Ceará},
                country={Brazil}}

\affiliation[2]{organization={Graduate Program in Teleinformatics Engineering (PPGETI), Federal University of Ceará (UFC)},
                addressline={Campus do Pici - Ac. Público, 725}, 
                city={Fortaleza}, 
                postcode={60455-970}, 
                state={Ceará},
                country={Brazil}}

\affiliation[3]{organization={Federal University of Rio de Janeiro (UFRJ)},
                addressline={R. Antonio Barros de Castro, 119}, 
                city={Rio de Janeiro}, 
                postcode={21941-853}, 
                state={Rio de Janeiro},
                country={Brazil}}

\received{1 May 2013}
\finalform{10 May 2013}
\accepted{13 May 2013}
\availableonline{15 May 2013}
\communicated{S. Sarkar}

\begin{abstract}

 We introduce a novel technique called \spnet\ to enhance body composition assessment.  This method employs a deep neural network capable of estimating Body Fat Percentage (BFP), performing individual identification, and enabling localization using a single photograph. The accuracy of \spnet\ is validated through comprehensive comparisons against the gold standard method, Dual-Energy X-ray Absorptiometry (DXA), utilizing 1273 healthy adults spanning various ages, sexes, and BFP levels. The results demonstrate that \spnet\ outperforms in 19.5\% state of the art computer vision-based approaches for body fat estimation, achieving a Mean Absolute Percentage Error (MAPE) of 4.91\% and Mean Absolute Error (MAE) of 1.42. The study evaluates both gender-based and Gender-neutral approaches, with the latter showcasing superior performance. The method estimates BFP with 95\% confidence within an error margin of 4.01\% to 5.81\%. This research advances multi-task learning and body composition assessment theory through \spnet.

\end{abstract}

\begin{keyword}
\MSC 41A05\sep 41A10\sep 65D05\sep 65D17
\KWD Keyword1\sep Keyword2\sep Keyword3

\end{keyword}

\end{frontmatter}


\section{Introduction}
\label{sec:introduction}



Article 25 of the Universal Declaration of Human Rights (1948) underscores the pivotal role of proper nutrition in maintaining good health. Malnutrition takes various forms, including wasting, stunting, micronutrient deficiencies, and diet-related non-communicable diseases \citep{WHO2021}. Global rise in obesity and overweight conditions is a significant challenge, characterized by excessive body fat accumulation, which poses a major health risk. The World Health Organization (WHO) reports a sharp increase in obesity prevalence since 1975, with over 1.9 billion adults and 340 million children and adolescents affected in 2016. Alarmingly, in many countries, health issues linked to obesity now surpass those related to being underweight, and obesity is associated with higher risks of chronic diseases like cancer, cardiovascular disease, and diabetes \citep{WHO2021}



It is alarming that in most countries, obesity and overweight conditions result in more fatalities than low body mass. These statistics demonstrate a urgent need for global action to prevent and manage obesity \citep{WHO2021}. Therefore, making assessments of body composition widely accessible becomes crucial. While various methods exist for estimating body composition, Dual-Energy X-ray Absorptiometry (DXA) remains the gold standard. DXA is a widely used reference technique for assessing body composition, including body fat, however, it is not universally available, particularly in urban centers, and tends to be more expensive than alternative diagnostic methods \citep{kuriyan2018body}.

Exploration of human body composition has a rich history spanning over 80 years. Over these decades, significant research milestones have emerged, with notable contributions from researchers \citep{siri1956gross, jackson1978generalized}. Despite these endeavors, a comprehensive understanding of the subject remains elusive. Presently, the scientific community advocates for methods characterized by enhanced accuracy, precision, and user-friendliness. Furthermore, Artificial Intelligence (AI) has witnessed rapid evolution for over six decades, and recent efforts have been dedicated to unlocking the potential of AI techniques in body composition estimation \citep{alves2021gender, majmudar2022smartphone, shao2014body}. This trend underscores a growing interest in this innovative approach.

Our objective is to develop a multi-task learning method using Convolutional Neural Networks (CNNs). This method aims to perform three simultaneous tasks: identifying individuals, pinpointing their coordinates within the image, and estimating their body fat percentage (BFP). Furthermore, the method is designed to emulate DXA exams, adhering to a quality benchmark of maintaining BFP estimation differences below 10\%. As a result, this study makes the following primary contributions:

\begin{enumerate}
    \item Accurate body fat percentage estimation: This research introduces a deep learning architecture for precise estimation of body fat percentage from a single photo.

    \item Validation and comparison to gold standard reference: The proposed method will be validated by comparing body fat percentage estimates to DXA.
    
    \item Expanded multi-task learning: Our research encompass tasks beyond body composition estimation. It enables simultaneous
    identification, localization, and regression.

    \item Novel Method Development: This research introduces an innovative method based on CNNs that not only estimates body fat percentage but also identifies individuals and estimate their body fatness.
\end{enumerate}


    


\section{Background and Related work}



The term Body Composition (BC) refers to the distribution and quantities of components within the total body mass \citep{holmes2021utility}. Historically, knowledge about human body composition relied on chemical analysis of organs and cadaver dissections, facilitating the quantification of fat, total water, minerals, and body protein content, which provided reference data for the development of body composition models.  The well known model has five levels: atomic, molecular, cellular, tissue, and whole body, with the total body mass accounting for the sum of components at each level \citep{wang1992five}.  The BC can be described as a two-compartment (2C) model, distinguishing fat mass and fat-free mass using methods like anthropometry or bioelectrical impedance analysis \citep{kuriyan2018body}. Additionally, a three-compartment model further divides fat-free mass into lean tissue mass and bone mineral content, with Dual-energy X-ray Absorptiometry (DXA) as an example method. DXA is also known as the golds standard method to estimate BC \citep{kuriyan2018body}. In summary, it's crucial to assess the agreement of DXA with contemporary body composition assessment methods to validate accuracy, identify biases, and ensure reliable measurement of body composition parameters.


 The literature review highlights the Visual Body Composition (VBC) model, which is an automated computer vision method using two smartphone photos to estimate total body fat percentage \citep{majmudar2022smartphone}. They used 134 healthy adults and DXA values as a references. The IRE-SVM, an improved machine learning approach that incorporates bias error control and feature selection to enhance prediction accuracy \citep{chiong2020using}. Hybrid models and regression techniques that yield more accurate predictions with fewer body measurements \citep{shao2014body}. Some relevant works employ hybrid machine learning methods and real anthropometric measurements to estimate body fat percentage \citep{alves2021gender, uccar2021estimation}.

Regarding body composition assessment theory, the
\spnet\ method is an innovative doubly indirect approach classified as a two-compartment method to estimate body fat percentage.

\section{Body Fat Estimation with ShapedNet}

The goal of our network is not limited to predict human's body fat percentage, but also localize and identify the person's body. 
The core idea is that the feature representations used to identify objects in inner layers empowers of meaningful representation to estimate body fat percentage as well. 

An input image containing a person is given to the model and after a series of operations, such as convolutions, the network provides the following outputs: (1) the body fat percentage; (2) the person's ROI information, \textit{i.e.} $x, y, w, h$; (3) person's object confidence and (4) the object class, in this case is set to 1, since the network is only trained to identify humans.


\subsection{Network Design}

\begin{figure*}[ht!]
	\centering
	 \includegraphics[width=.7\linewidth]{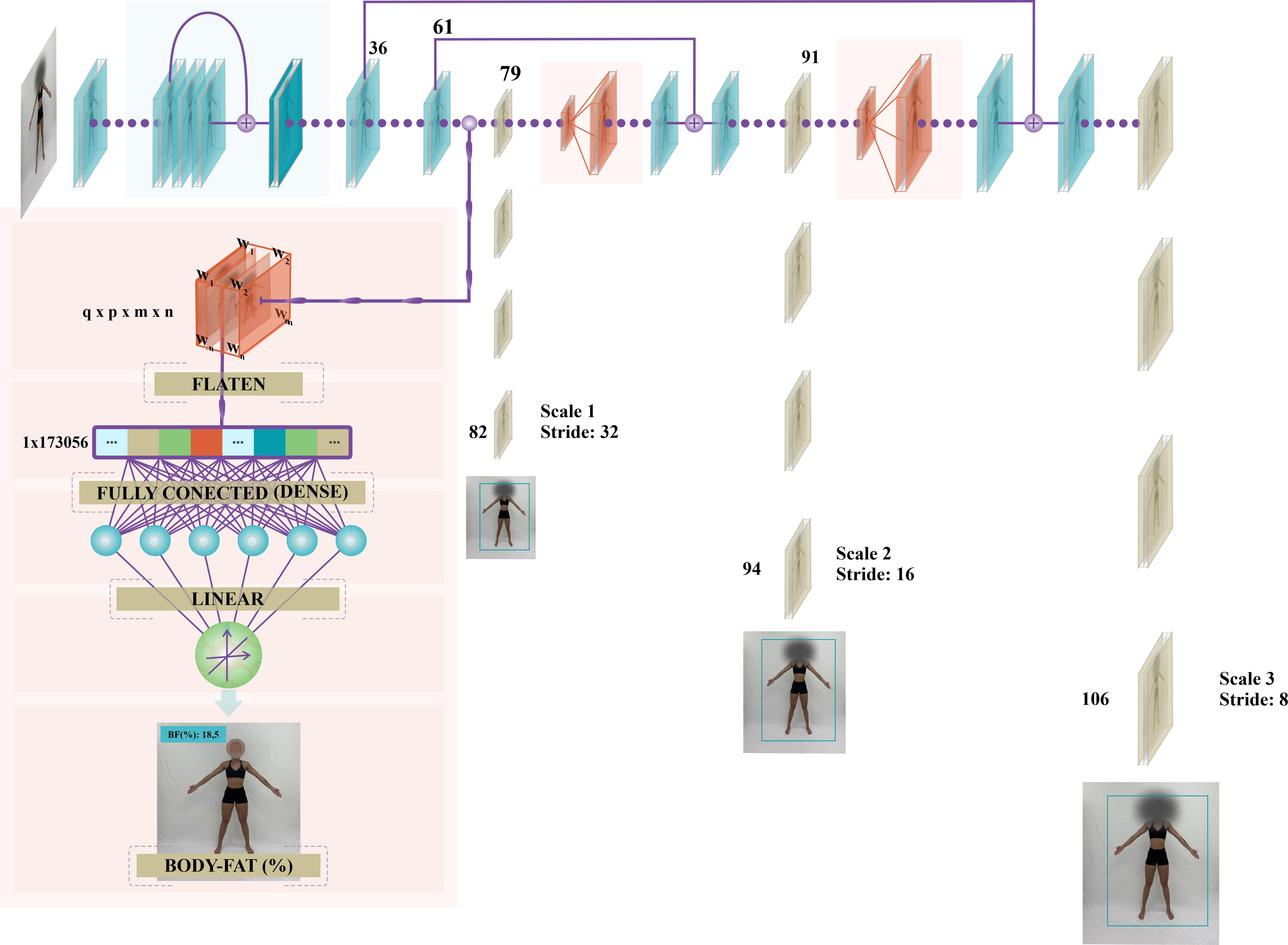}
	 \caption{The higher-level architecture of ShapedNet.}
	 \label{fig:met-ShapedNet-02}
\end{figure*}

We changed a design found in literature to develop our architecture \citep{YOLOv3}. The higher level architecture of the \spnet\ is summarized in Fig. \ref{fig:met-ShapedNet-02}. The upper part of the figure are a series of convolutional layers, residual, concatenation and stride, but most of it is ommited in this figure and will be explained later. The layer's numbers are the top part of the figure and the first 53 layers are built based on a popular architecture, name Darknet-53. And, right after the last layer of the network we created a branch, which is highlighted in the figure. The input of the branch is modeled as a 4-D tensor, with dimensions $q\times p \times m \times n$, wherein $q=1024$, $p=13$, $m=13$, $n=1$. Thus, a flatten layer vectorizes everything into a 1-D tensor of size $1 \times 173056$, then  the tensor flows to a fully connected layer, \textit{i.e.} Dense with 173056 neurons and 1 output neuron using a linear function to estimate body fat percentage.
Another outpus are created from layers 79, 91 and 103, these were originally developed to make the model able to perform a multi scale detection \citep{YOLOv3}, which means detect objects at different scales and sizes.

It is important to note that operations such as batch normalization, dropout, activation function are still present, they were only omitted for simplicity. There are three layers here, one flatten, one dense with a linear activation function, which is represented as another layer in this illustration just to the reader should keep in mind that others functions commonly used in regression could be employed as well.


Therefore, there are three main parts that composes the object detection and classification outputs of the ShapedNet model. They are $\hat{\vec{Y}}_{loc}$, the one that will be later used to estimate coordinates of the given person, which is an tensor of size $ [ bs \times cs \times cs \times 3 \times 2 ]$. Plus, $\hat{\vec{Y}}_{conf}$, that estimates the confidence of the prediction bounding box, this is a tensor of size $ [ bs \times cs \times cs \times 3 \times 2 ]$ as well. And, $\hat{\vec{P}}_{c}$, which gives the probability distribution of the given classes, in a tensor size $ [ bs \times cs \times cs \times 3 \times nc ]$. So, these ouputs composes a multi dimensional tensor, represented as follows,

\begin{equation}
	\label{eq:met:tensor_1}
	\begin{aligned}
		\hat{\vec{Y}}_{loc} = 
		\left[\begin{array}{cccc}
		\hat{\vec{Y}}_{loc} & \hat{\vec{Y}}_{conf} & \hat{\vec{P}}_{c},
		\end{array} 
		\right].
	\end{aligned}
\end{equation}

Additionaly, the output used to estimate body fat percentage is $\hat{\vec{Y}}_{bf}$ of ShapedNet is, 

\begin{equation}
	\begin{aligned}
		\hat{\vec{Y}}_{bf} = f(\vec{y_{fl}}^T \vec{w_{bf}}),
	\end{aligned}
\end{equation}

\noindent being, $\vec{w_{bf}}$ the vector of parameters that ought to be optimize later. Its size is $[1 \times 173056]$. And, 

\begin{equation}
	\begin{aligned}
		\vec{y_{fl}} = vec(\vec{y_{conv-53}}),
	\end{aligned}
\end{equation}

\noindent wherein $vec(\cdot)$ is the vectorization function, that transforms a tensor of any size into a 1-D tensor. And $f$ is the identify function, $f(X) = X$. 


Let the output tensor of Equation \ref{eq:met:tensor_1} be expanded to 

\begin{equation}
	\label{eq:met:tensor_2}
	\begin{aligned}
		\hat{\vec{Y}}_{loc} = 
		\left[\begin{array}{cc|cc}
		\hat{\vec{Y}}_{loc} & \hat{\vec{Y}}_{conf} & \hat{\vec{P}}_{c} & \hat{\vec{Y}}_{bf},
		\end{array} 
		\right],
	\end{aligned}
\end{equation}

\noindent and now we have the full output of the ShapedNet.

The addition of the term $\hat{\vec{Y}}_{bf}$ to $\vec{Y}$ requires that a loss term is considered to optimize the parameters in $\vec{w_{bf}}$. So, a new term, ${\large \textit{L}_f}$, was added to the original loss function \citep{YOLOv3}. So, we rewrite the following multi-part loss function:

\begin{equation}
	\begin{aligned}
		{\large \textbf{L}} \  = \ &  \ \lambda_\textbf{coord}
		\sum_{i = 0}^{S^2}
			\sum_{j = 0}^{B}
			 \mathlarger{\mathbb{1}}_{ij}^{\text{obj}}
					\left[
					\left(
						x_i - \hat{x}_i
					\right)^2 +
					\left(
						y_i - \hat{y}_i
					\right)^2 
					\right]
					\\
			& + \ \lambda_\textbf{coord} 
			\sum_{i = 0}^{S^2}
				\sum_{j = 0}^{B}
					 \mathlarger{\mathbb{1}}_{ij}^{\text{obj}} \times 
                &
                \\
                & 
					 \left[
					\left(
						\sqrt{w_i} - \sqrt{\hat{w}_i}
					\right)^2 +
					\left(
						\sqrt{h_i} - \sqrt{\hat{h}_i}
					\right)^2
					\right]
			\\
			& + \ \sum_{i = 0}^{S^2}
			\sum_{j = 0}^{B}
				\mathlarger{\mathbb{1}}_{ij}^{\text{obj}}
				\left(
					C_i - \hat{C}_i
				\right)^2 
			\\
			& + \ \lambda_\textrm{no-obj}
			\sum_{i = 0}^{S^2}
				\sum_{j = 0}^{B}
				\mathlarger{\mathbb{1}}_{ij}^{\text{no-obj}}
					\left(
						C_i - \hat{C}_i
					\right)^2
			\\
			& + \ \sum_{i = 0}^{S^2}
			\mathlarger{\mathbb{1}}_i^{\text{obj}}
				\sum_{c \in \textrm{classes}}
					\left(
						p_i(c) - \hat{p}_i(c)
					\right)^2
			\\
			& + \ { \Large \textit{L}_f }, 
	\end{aligned}
\end{equation}

\noindent The ${\large \textit{L}_f }$ is the loss term to measure body fat percentage error, and is expressed as, 

\begin{equation}
	{ \Large \textit{L}_f }  = \lambda_\textbf{f} \ \frac{(b - \hat{b})}{b} .
\end{equation}


\noindent Note that $b$ is the true value of body fat percentage and $\hat{b}$ is the predicted value. $\lambda_\textbf{f}$ is a value used to provide a weight, usually equal to $1$, to the total loss. 


\noindent We have incorporate the idea of finetuning, and used the original weights of the pre-trained as a starting point during training.




The $\vec{w_{bf}}$, \textit{i.e.} the weights connected to the new regression output $\hat{\vec{Y}}_{bf}$ are initialized using as recommendations found on literature \citep{pmlr-v9-glorot10a}. We have used learning rate warm up \citep{goyal2017accurate}. In our case, the first 2 epochs are used for warm up and the initial learning is $\eta_i=1\mathrm{e}^{-4}$. 








And a cosine learning rate decay is used after warm up procedure \citep{loshchilov2016sgdr}, because it has demonstrated to enhance accuracy and led to better transfer learning performance in other applications such as semantic segmentation. As optimizer we have used ADAM \citep{kingma2014adam}, which is widely know and used in deep learning applications. The criteria used to stop training is the number of epochs. When the training reaches 1000 epochs it stops and the chosen weights are the one that reaches the lowest loss in the validation dataset.

\subsection{Database and Experiments}
\label{sec:met:dataset}

Creating databases for body composition assessments is a costly and time-consuming process. In previous studies \citep{uccar2021estimation, chiong2021using, shao2014body}, researchers have reported the use of public databases as an alternative. One commonly used resource is the StatLib database, which contains anthropometric measurements for 252 men, including age, weight, height, and ten body circumferences. However, it should be noted that this dataset does not include data for female individuals. Furthermore, the StatLib database uses an equation \citep{siri1956gross} as the reference method for body fat estimation, rather than DXA, and does not provide photographs of the subjects.

To address this gap, we conducted a comprehensive data collection effort involving anthropometric measures, photographs, and DXA exams from both male and female individuals. This collaborative effort involved research groups from the Federal University of Rio de Janeiro (UFRJ) and the Federal Institute of Ceará (IFCE). The clinical trials and examinations took place at the Josue Carlos Nutrition Institute at UFRJ, while the analysis system was implemented at IFCE. Ethical compliance was ensured by registering the dataset under the National Health Council through our university, following the necessary procedures for research involving human subjects. The study adhered to the guidelines and requirements of the Research Ethics Committee of the IFCE, protocol number 3593367. The project is registered in Plataforma Brasil under CAAE registration 16885519.1.0000.5589. Participants provided their informed consent for the use of the collected data in research, confirming compliance with the required parameters.

Our dataset is named \bcdb, it consists of 1273 subjects, being 578 men and 695 women. It includes information such as age, height, weight and ten anthropometric measures. The images were obtained using a smartphone camera with a resolution of 1920 × 1080 pixels and 8-bit depth. Each individual's body was captured from frontal, back, right-side, and left-side perspectives. The dataset covers individuals aged between 18 and 65, with DXA values ranging from 9.30 to 57.60.

\subsection{Evaluation Metrics and Hypothesis Test}

In our experiments, we allocated 10\% of the subjects for testing (127), 10\% for validation (127), and 80\% for training purposes (1019). The validation set was specifically used for early stopping during the training process. To ensure representative sampling, the subjects were stratified based on their Body Mass Index (BMI), a commonly used indicator to assess the relationship between body weight and height. The dataset was divided in a way that maintained the proportions of the six BMI categories across the data subsets. 

The metrics evaluate are the Mean Absolute Percentage Error (MAPE), Mean Absolute Error (MAE) and Mean Relative Error (MRE).  Note that the MAPE gives a better representation than MAE, because it percentage metric relative to the true value. While MAE provides the real difference in measurements. This is particularly important, considering women tend to have a higher BFP than in men. Moreover, the MRE shows how the error is displaced around zero, which means is this metrics provides an indicative if the model is underestimating or overestimating the body fat percentage. Our measure of success are errors below 10\%. 

Moreover, we also report the Confidence Interval (CI) at 95\% level, for each one of the metris above. In our case of estimating the mean of a population using a \textit{t}-distribution with a given confidence level (95\% or $\alpha=0.05$). CI values are reported as $[CI_{min}, CI_{max}]$.

We also perform Tukey's HSD test as a \textit{post-hoc} statistical test used to compare multiple group means. The confidence is set to 95\%, so for $p<.05$ there are enough evidences to reject the null hypothesis, $\vec{H_0}$, and accept the alternative hypothesis, $\vec{H_1}$. For $p\geq.05$ there are evidences to accept the null hypothesis $\vec{H_0}$. We address the following question in our research: \textbf{Is a gender-based approach better than a Gender-neutral based one?} Then, we design the hypothesis:

\begin{itemize}
	\item[$\vec{H_0}$:] No significant differences are observed when using a gender-based approach or a Gender-neutral-based approach for \spnet.
 
	\item[$\vec{H_1}$:] Using a gender-based approach is in fact different than using a Gender-neutral one to train the \spnet.
\end{itemize}

\section{Results and Discussions}

We present a analysis of the \spnet, starting with a quantitative assessment of its accuracy. The first part focus on evaluating the error against DXA as the gold standard, providing insights into the method's performance. Then, we test how different gender-based and  Gender-neutral based approaches are. Finally, the sections concludes with a comparison of \spnet to state-of-the-art (SOTA) methods, highlighting its relative performance in the field.

\subsection{Accuracy Assessment: Evaluating Error against DXA}

We evaluate our method on training, validation and test datasets, as shown in Tab. \ref{tab:results:ShapedNet:overall_results}. The reported values are mean and standard deviation of MAE, MAPE and MSD in respect to DXA.  The results for the training and validation sets are provided to assess the model's performance and mitigate overfitting. Each row in the table represents a different model training. The models "Male" and "Female" are gender-based models trained using data from individuals of the corresponding gender. Conversely, the Gender-neutral model is not designed to consider gender-specific information, making it a more general approach.

\spnet\ achieved a mean absolute error (MAE) in body fat percentage of 1.59 for males and 2.03 for females. The corresponding MAPE values indicate that, on average, the measurement error accounted for 7.04\% and 5.98\% of the total body fat estimation for males and females, respectively. Notably, the Gender-neutral model outperformed the gender-specific models, demonstrating an MAPE of 4.91\% (1.42) in error.

The MAE values in the test set are comparable to those in the validation set, indicating that the model did not overfit during training. For example, the MAE for males was 1.59 in the test set and 1.73 in the validation set, while for females it was 2.03 in the test set and 2.25 in the validation set. The Gender-neutral model exhibited a slightly larger difference between the sets, particularly in the train set. The average error during training for this model was 0.14, however it is not and indication of overfitting since errors in validation (1.51) and test (1.42) sets are similar. This is actually promising, as it suggests that a single model performs better than gender-specialized models.

However, the reported values of MSD are interesting. The female model tends to undermeasure the body fat percentage in women. For instance, the model tends to estimate the body fat percentage -0.21, on average, in women. And this behavior happens in all sets of data. This might be due to factors such as body fatness density location in women are different than in men. For example, is well-know that women store more fat in the gluteal-femoral region than men and we are using only the front photo to estimate BFP, so it is possible that the model is not considering different angles of the person. And, this is an investigation that ought to be addressed in the future.

It is important to pay attention to the confidence intervals presented in Tab. \ref{tab:ShapedNet:results_CI}. These intervals were estimated with a significance level of $\alpha = 0.05$, indicating that for 95\% of the subjects, the error in body fat percentage estimation lies between 5.89\% and 8.19\% for the male group. Another important consideration is the spread, denoted as $\upsilon = CI_{\text{max}} - CI_{\text{min}}$. A narrower spread indicates less uncertainty in our estimates. For the male, female, and Gender-neutral models, the spreads are $\upsilon_{m} = 2.3\%$, $\upsilon_{f} = 1.72\%$, and $\upsilon_{ng} = 1.8\%$, respectively.

The negative mean signed difference values observed in the female group are further supported by the confidence intervals presented in Tab. \ref{tab:ShapedNet:results_CI}. The lower bound value of -6.53\% indicates that the error in the female group is predominantly negative, suggesting that the estimated body fat percentage using ShapedNet is consistently lower than the DXA measurements for females. This reaffirms that the female BFP is being underestimated compared to DXA.

Let us discuss if a gender is relevant to the model's performance. A gender-neural model appears to outperform gender-based approaches, suggesting that a single model could be utilized instead of two during production, resulting in a more efficient system. This finding contrasts with other gender-based approaches that have been traditionally used in clinical procedures. However, it is crucial to determine if these approaches are indeed different. To investigate this, a Tukey's HSD paired test was conducted, and the results are presented in Tab. \ref{tab:ShapedNet:results_tukey}. A significance level of $p < 0.05$ was chosen to reject the null hypothesis, indicating the presence of statistically significant differences among the groups. The absolute error values were used as variables in the tests, enhancing their statistical power.

As a result, the null hypothesis $\vec{H_0}$ is rejected, indicating that the Gender-neural model's estimations are indeed different from those of the male or female models in terms of BFP calculations. Consequently, it is possible to conclude that the gender-neural model has a lower mean absolute error compared to using a single model, with a difference of 0.17 ($1.59 - 1.42$) for males and 0.61 ($2.03 - 1.42$) for females. Additionally, the percentage error of the gender-neural model is 2.13 points lower for males ($7.04\% - 4.91\%$) and 1.07\% lower for females ($5.98\% - 4.91\%$) compared to gender-based models.

\begin{table}[H]
  \caption{Overall results on test, validation, and train sets. The metrics are MAPE, MAE, and MSD, and the groups of analyses are Male, Female, and Gender-neural.}
  \centering
  \begin{tabular}{c|c|c|c}
  \toprule
   \textbf{Gender/Metric} & \textbf{MAPE (\%)}                                                     & \textbf{MAE}        &\textbf{MSD} \\
   \midrule \midrule
  \multicolumn{4}{c}{\textbf{Test}}                                                                                            \\
                \midrule \midrule
  Male          & 7.04 $\pm$ 6.43                                            & 1.59 $\pm$ 1.38    &  0.38 $\pm$ 2.08                                            \\
  Female        & 5.98 $\pm$ 5.05                                            & 2.03 $\pm$ 1.58           &  -0.21 $\pm$ 2.57                                   \\
  Gender-neural     & \textbf{4.91 $\pm$ 6.78}                                           & \textbf{1.42 $\pm$ 1.82}                        &  \textbf{0.28 $\pm$ 2.29}                      \\
              \midrule \midrule
  \multicolumn{4}{c}{\textbf{Validation}}                                                                                             \\
                \midrule \midrule
  Male          & 7.23 $\pm$ 6.50                                           & 1.73 $\pm$ 1.52     &  0.45 $\pm$ 2.26                                         \\
  Female        & 6.56 $\pm$ 6.46                                            & 2.25 $\pm$ 1.92                                    &  -0.08 $\pm$ 2.96          \\
  Gender-neural     & 5.15 $\pm$ 6.01                                           & 1.51 $\pm$ 1.67                              &  0.05 $\pm$ 0.06                 \\
              \midrule \midrule
  \multicolumn{4}{c}{\textbf{Train}}                                                                                           \\
                \midrule \midrule
  Male          & 4.64 $\pm$ 4.21                                            & 1.10 $\pm$ 0.97                 &  0.28 $\pm$ 1.44                             \\
  Female        & 3.90 $\pm$ 3.36 & 1.35 $\pm$ 1.06 &  -0.38 $\pm$ 1.68  \\
  Gender-neural     & 0.52 $\pm$ 2.01                                           & 0.14 $\pm$ 0.48  &  0.03 $\pm$ 0.50 \\
  \bottomrule
  \end{tabular}
  \label{tab:results:ShapedNet:overall_results}
  \end{table}

  \begin{table}[H]
\centering
\caption{Error Analysis with 95\% Confidence Intervals: MAPE, MAE, and MSD.}
\begin{tabular}{c|ccc}
\toprule
          & \textbf{MAPE (\%)}              & \textbf{MAE}     & \textbf{MSD}        \\
          \midrule
\textbf{Male}      & {[}5.89, 8.19{]}  & {[}1.34, 1.84{]} & {[}0.01, 0.75{]}\\
\textbf{Female}    & {[}5.12, 6.84{]}   & {[}1.76, 2.30{]} & {[}-0.65, 0.22{]}\\
\textbf{Gender-neural} & {[}4.01, 5.81{]} & {[}1.18, 1.66{]} & {[}-0.01, 0.59{]} \\
\bottomrule
\end{tabular}
\label{tab:ShapedNet:results_CI}
\end{table}

\begin{table}[H]
\centering
\caption{Results of the Tukey's HSD hyphotesis test. MD denotes Mean Differences and the values of $p$ highlighted in bold represent the groups are significantly different at the 95\% confidence level}
\begin{tabular}{c|cc|cc}
\toprule
\textbf{Variable}              & \multicolumn{2}{c}{\textbf{Groups}} & \textbf{MD}      & $\textbf{p}$   \\
\midrule
\multirow{2}{*}{\textbf{MAE}}  & Gender-neural     & Male       & -1.0273 & \textbf{.001}  \\
                      & Gender-neural     & Female     & -1.4612 & \textbf{.001} \\
                      \bottomrule
\end{tabular}
\label{tab:ShapedNet:results_tukey}
\end{table}

We shown \spnet\ exam examples from six subjects in Fig. \ref{fig:results_in_subjects}, deliberately selecting individuals of both sexes and representing three distinct biotypes: ectomorph, mesomorph, and endomorph. The error displayed in the figure represents the relative difference between DXA and \spnet\ measurements. Our aim is to exhibit the accuracy of our method across different body types.

\subsection{Comparing with State-of-the-Art (SOTA) Methods}
  
A comparison with state-of-the-art (SOTA) approaches for estimating body fat percentage is shown in Tab. \ref{tab:results_comparison}. The table is presented in descending order, with the most recent approach from the literature at the top. "N" represents the number of subjects used in the study, while "RM" refers to the reference method used for body fat estimation. The values of MAE and MAPE for males, females, and individuals of gender-neural are reported for comparison.

\begin{table*}[!ht]
    \caption{Comparative Analysis: \spnet\ vs. SOTA Approaches for BFP estimate. Metrics: MAE, MAPE for Male, Female, Non-Gender. N: Number of Subjects. RM: Reference Method. Dash indicates missing information.}
    \adjustbox{width=\textwidth}{
    \begin{tabular}{cc cc cc cc}
    \toprule
    \textbf{Approach} & \textbf{Author} & \textbf{Input} & \textbf{N} & \textbf{RM} & \textbf{MAE\textsubscript{m} / MAPE\textsubscript{m}}& \textbf{MAE\textsubscript{f} / MAPE\textsubscript{f}} & \textbf{MAE\textsubscript{ng} / MAPE\textsubscript{ng}} \\
    \midrule
    \spnet & Our approach & Front photo & 1273 & DXA & 1.59 / 7.04\% & 2.03 / 5.98\% & \textbf{1.42 / 4.91\%} \\
    \midrule
    VBC & \citep{majmudar2022smartphone} & \begin{tabular}[c]{@{}c@{}}Front photo\\ +\\ Back photo\end{tabular} & 134 & DXA & 1.88 / 6.8\% & 2.43 / 6.1\% & \textemdash \\
    \hline
    L-SVM & \citep{Shara2022} & \begin{tabular}[c]{@{}c@{}}6 CV-AM\\ +\\ IV\end{tabular} & 912 & DXA & 3.24 / 15\% & 3.44 / 11\% & 3.47 / 13\% \\
    \hline
    LSSVR & \citep{alves2021gender} & \begin{tabular}[c]{@{}c@{}}10 AM\\ +\\ IV\end{tabular} & 163 & DXA & 2.75 / \textemdash & 4.01 / \textemdash & 3.56 / \textemdash \\
    \hline
    DT-SVM & \citep{uccar2021estimation} & \begin{tabular}[c]{@{}c@{}}13 AM\end{tabular} & 252 & Siri-BD & 3.53 / 30.06\% & \textemdash & \textemdash \\
    \hline
    IRE-SVM & \citep{chiong2021using} & \begin{tabular}[c]{@{}c@{}}13 AM\end{tabular} & 252 & Siri-BD & 3.62 / \textemdash & \textemdash & \textemdash \\
    \hline
    MR-MARS & \citep{shao2014body} & \begin{tabular}[c]{@{}c@{}}13 AM\end{tabular} & 252 & Siri-BD & 3.69 / 24.28\% & \textemdash & \textemdash \\
    \bottomrule
    \end{tabular}
    }
    \label{tab:results_comparison}
\end{table*}

Our approach, \spnet, utilizes front photos while the VBC \citep{majmudar2022smartphone} uses two photos as inputs. Both \spnet\ and VBC uses DXA as the reference method. The method developed by these researchers, the VBC, is the most similar to ours as it also employs CNN-based models. The results achieved by the researches  are considered relevant, as they achieved a MAPE of less than or equal to 10\% \citep{majmudar2022smartphone}. However, the number of subjects in their study represents only 10.5\% of our dataset. Our study is more robust in terms of generalization, compared to them \citep{majmudar2022smartphone}, due to the inclusion of a significantly larger sample size, with 9.5 times more human subjects. Additionally, a key advantage of \spnet\ over VBC is that our approach only requires one photo, whereas VBC requires two photos (front and back views). At last but not least, VBC requires preprocessing in the input images before feeding VBC, such as background removal using another CNN model, while \spnet\ does not require any preprocessing.

We achieved an MAE of 1.59 and MAPE of 7.04\% for male subjects, and an MAE of 2.03 and MAPE of 5.98\% for female subjects. In comparison, \citep{majmudar2022smartphone} achieved similar results, but in a much smaller sample size. The notable difference lies in the gender-neural  model, where we achieved an MAE of 1.42 and MAPE of 4.91\%. This represents an improvement of 1.19 to 1.89 points in MAPE for male and female subjects. Moreover, the resulted reported by the authors are not clear if are in the test dataset or in the all subjects.

The proposed method demonstrates a significantly lower error of 1.65 compared to for male subjects \citep{Shara2022}. Furthermore, the error is 1.16, 2.03,2.10 and 1.94 lower than the methods proposed by \citep{alves2021gender, chiong2021using, shao2014body, uccar2021estimation} respectively. It is important to note that these methods are restricted to the male gender. In contrast, our study encompasses a broader analysis, with a dataset of 1273 individuals, including 578 men and 695 women. This significantly larger study size allows for more robust conclusions, particularly for the male population, which is approximately twice the size of the populations in the previously mentioned studies.

The only authors that studied female subjects were \citep{majmudar2022smartphone}, \citep{Shara2022} and \citep{alves2021gender}. Regarding their results, the error in our method is 0.43 lower than \citep{majmudar2022smartphone}, and 1.41 and 1.98 for \citep{Shara2022} and \citep{alves2021gender} respectively. Among the authors presented in Tab. \ref{tab:results_comparison} only \citep{majmudar2022smartphone}, \citep{Shara2022}, and \citep{alves2021gender} studied female subjects. In terms of results, our method achieves an error in BFP that is 0.43 lower than \citep{majmudar2022smartphone}, and 1.41 and 1.98 lower than \citep{Shara2022} and \citep{alves2021gender}, respectively.

According to the baseline metrics models that have MAPE errors above 10\% are considered inappropriate for use. Therefore, the errors reported by \citep{shao2014body} and \citep{chiong2021using}, which are 30.06\% and 25.28\%, respectively, are significantly higher compared to our method, which achieves an error of 4.91\%. In these studies, the errors are 2-3 times larger than our method.

\def\COL_PROPORTION_2{1}
\begin{figure}[!ht]
    \centering
    \includegraphics[trim={0cm 0cm 9.5cm 0cm}, clip, page=9, width=\COL_PROPORTION_2\columnwidth]{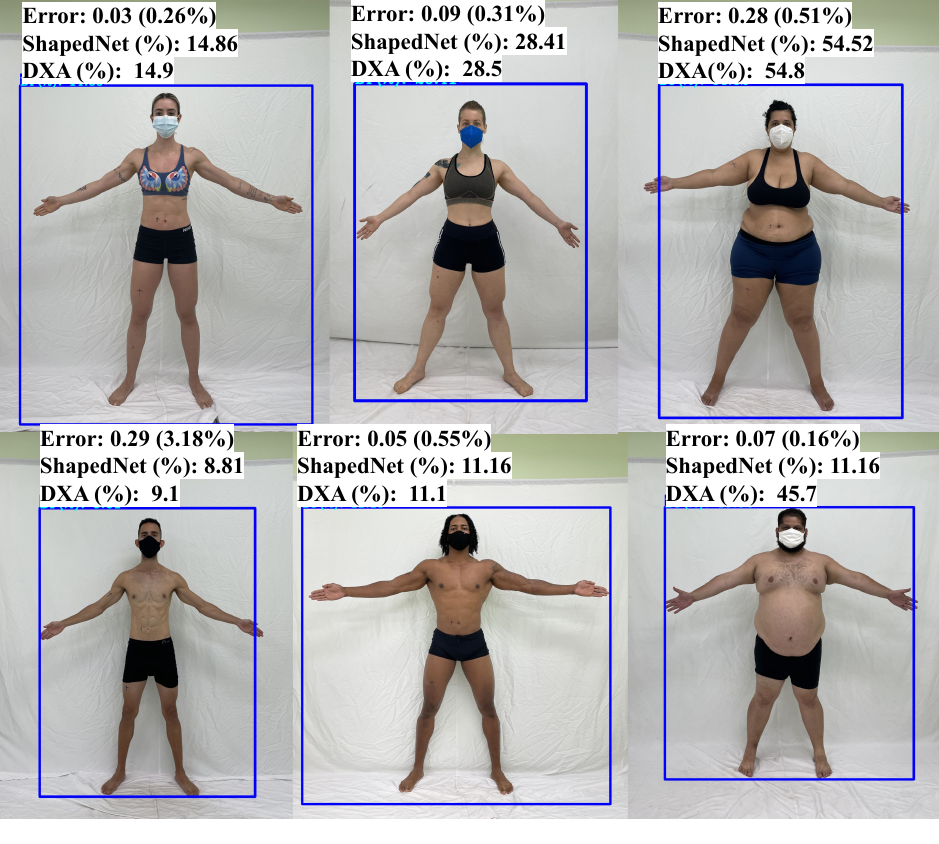}
    
    \caption{Results of the Gender-neutral \spnet\ model for both female and male subjects are presented, along with estimates from DXA. The displayed error values are in comparison to DXA measurements.}
    
    \label{fig:results_in_subjects}
\end{figure}

\subsection{Limitations of \spnet}

One limitation of \spnet\ is related to the input photos used during data acquisition. The photos were taken in a controlled environment with a white background, and the network was trained using these specific background conditions. However, in clinical practice, photos can be captured in various backgrounds and scenes, which may negatively impact the performance of the method. Although some preliminary tests were conducted using individuals outside of the original study in different scenarios, the results were inconclusive. While the estimation of body fat percentage did not seem to be significantly affected, there were challenges associated with person identification and localization. Therefore, further investigations are required to gain a better understanding of this issue and develop strategies to address it effectively.

We did not perform a dedicated investigation or quantification of how camera angle and scene illumination specifically impact the estimation of body fat percentage. However, during the data collection process, we made slight adjustments to the camera position by varying the angles from up and down. This was done in order to capture potential variations caused by the camera's angle and incorporate them into the dataset. While this approach aimed to account for some variations, further studies focusing specifically on the impact of camera angle and scene illumination on BFP estimation would provide valuable insights and help enhance the robustness of the method.

\section{Conclusions}

Our study represents a significant advancement in accurate body fat estimation through advanced computer vision techniques. Our method not only surpasses previous state-of-the-art approaches in terms of accuracy and generalization but also introduces a simpler technique. With just a single photo, our approach demonstrates a 19.5\% improvement in accuracy for estimating body fat percentage. The Gender-neutral model consistently outperforms the best reported results for computer vision-based body fat estimation using photographs, with a margin ranging from 1.19 to 1.89 points.

The validation against the gold standard, Dual-Energy X-ray Absorptiometry (DXA), establishes a Mean Absolute Percentage Error (MAPE) of 4.91\% and a Mean Absolute Error (MAE) of 1.42. Our hypothesis testing underscores the superiority of the Gender-neutral model, surpassing the female-based model by 1.07 points and the male-based model by 2.12 points. In terms of precision, ShapedNet provides estimations within a narrow range of 4.01\% to 5.81\% compared to DXA, showcasing its reliability. Our research significantly advances accurate body fat estimation, with implications not only for body composition assessment but also for broader applications across diverse domains.

\bibliographystyle{model5-names}
\bibliography{refs}

\begin{thebibliography}{18}
\expandafter\ifx\csname natexlab\endcsname\relax\def\natexlab#1{#1}\fi
\providecommand{\url}[1]{\texttt{#1}}
\providecommand{\href}[2]{#2}
\providecommand{\path}[1]{#1}
\providecommand{\DOIprefix}{doi:}
\providecommand{\ArXivprefix}{arXiv:}
\providecommand{\URLprefix}{URL: }
\providecommand{\Pubmedprefix}{pmid:}
\providecommand{\doi}[1]{\href{https://doi.org/#1}{\path{#1}}}
\providecommand{\Pubmed}[1]{\href{pmid:#1}{\path{#1}}}
\providecommand{\bibinfo}[2]{#2}
\ifx\xfnm\relax \def\xfnm[#1]{\unskip,\space#1}\fi
\bibitem[{Alves et~al.(2021)Alves, Ohata, Nascimento, De~Souza, Holanda,
  Loureiro \& Rebou{\c{c}}as~Filho}]{alves2021gender}
\bibinfo{author}{Alves, S.~S.}, \bibinfo{author}{Ohata, E.~F.},
  \bibinfo{author}{Nascimento, N.~M.}, \bibinfo{author}{De~Souza, J.~W.},
  \bibinfo{author}{Holanda, G.~B.}, \bibinfo{author}{Loureiro, L.~L.}, \&
  \bibinfo{author}{Rebou{\c{c}}as~Filho, P.~P.} (\bibinfo{year}{2021}).
\newblock \bibinfo{title}{Gender-based approach to estimate the human body fat
  percentage using machine learning}.
\newblock In {\it \bibinfo{booktitle}{2021 International Joint Conference on
  Neural Networks (IJCNN)}\/} (pp. \bibinfo{pages}{1--8}).
\newblock \bibinfo{organization}{IEEE}.
\bibitem[{Alves(2022)}]{Shara2022}
\bibinfo{author}{Alves, S. S.~A.} (\bibinfo{year}{2022}).
\newblock {\it \bibinfo{title}{Sex-based approach to estimate human body fat
  percentage from 2D camera images with Deep Learning and Machine Learning}\/}.
\newblock Ph.D. thesis Universidade Federal do Ceara.
\bibitem[{Chiong et~al.(2020)Chiong, Fan, Hu \& Chiong}]{chiong2020using}
\bibinfo{author}{Chiong, R.}, \bibinfo{author}{Fan, Z.}, \bibinfo{author}{Hu,
  Z.}, \& \bibinfo{author}{Chiong, F.} (\bibinfo{year}{2020}).
\newblock \bibinfo{title}{Using an improved relative error support vector
  machine for body fat prediction}.
\newblock {\it \bibinfo{journal}{Computer Methods and Programs in
  Biomedicine}\/},  {\it \bibinfo{volume}{198}\/}, \bibinfo{pages}{105749}.
\bibitem[{Chiong et~al.(2021)Chiong, Fan, Hu \& Chiong}]{chiong2021using}
\bibinfo{author}{Chiong, R.}, \bibinfo{author}{Fan, Z.}, \bibinfo{author}{Hu,
  Z.}, \& \bibinfo{author}{Chiong, F.} (\bibinfo{year}{2021}).
\newblock \bibinfo{title}{Using an improved relative error support vector
  machine for body fat prediction}.
\newblock {\it \bibinfo{journal}{Computer Methods and Programs in
  Biomedicine}\/},  {\it \bibinfo{volume}{198}\/}, \bibinfo{pages}{105749}.
\bibitem[{Glorot \& Bengio(2010)}]{pmlr-v9-glorot10a}
\bibinfo{author}{Glorot, X.}, \& \bibinfo{author}{Bengio, Y.}
  (\bibinfo{year}{2010}).
\newblock \bibinfo{title}{Understanding the difficulty of training deep
  feedforward neural networks}.
\newblock In \bibinfo{editor}{Y.~W. Teh}, \& \bibinfo{editor}{M.~Titterington}
  (Eds.), {\it \bibinfo{booktitle}{Proceedings of the Thirteenth International
  Conference on Artificial Intelligence and Statistics}\/} (pp.
  \bibinfo{pages}{249--256}).
\newblock \bibinfo{address}{Chia Laguna Resort, Sardinia, Italy}:
  \bibinfo{publisher}{PMLR} volume~\bibinfo{volume}{9} of {\it
  \bibinfo{series}{Proceedings of Machine Learning Research}\/}.
\newblock \URLprefix \url{https://proceedings.mlr.press/v9/glorot10a.html}.
\bibitem[{Goyal et~al.(2017)Goyal, Doll{\'a}r, Girshick, Noordhuis, Wesolowski,
  Kyrola, Tulloch, Jia \& He}]{goyal2017accurate}
\bibinfo{author}{Goyal, P.}, \bibinfo{author}{Doll{\'a}r, P.},
  \bibinfo{author}{Girshick, R.}, \bibinfo{author}{Noordhuis, P.},
  \bibinfo{author}{Wesolowski, L.}, \bibinfo{author}{Kyrola, A.},
  \bibinfo{author}{Tulloch, A.}, \bibinfo{author}{Jia, Y.}, \&
  \bibinfo{author}{He, K.} (\bibinfo{year}{2017}).
\newblock \bibinfo{title}{Accurate, large minibatch sgd: Training imagenet in 1
  hour}.
\newblock {\it \bibinfo{journal}{arXiv preprint arXiv:1706.02677}\/}, .
\bibitem[{Holmes \& Racette(2021)}]{holmes2021utility}
\bibinfo{author}{Holmes, C.~J.}, \& \bibinfo{author}{Racette, S.~B.}
  (\bibinfo{year}{2021}).
\newblock \bibinfo{title}{The utility of body composition assessment in
  nutrition and clinical practice: an overview of current methodology}.
\newblock {\it \bibinfo{journal}{Nutrients}\/},  {\it \bibinfo{volume}{13}\/},
  \bibinfo{pages}{2493}.
\bibitem[{Jackson \& Pollock(1978)}]{jackson1978generalized}
\bibinfo{author}{Jackson, A.~S.}, \& \bibinfo{author}{Pollock, M.~L.}
  (\bibinfo{year}{1978}).
\newblock \bibinfo{title}{Generalized equations for predicting body density of
  men}.
\newblock {\it \bibinfo{journal}{British Journal of Nutrition}\/},  {\it
  \bibinfo{volume}{40}\/}, \bibinfo{pages}{497--504}.
\bibitem[{Kingma \& Ba(2014)}]{kingma2014adam}
\bibinfo{author}{Kingma, D.~P.}, \& \bibinfo{author}{Ba, J.}
  (\bibinfo{year}{2014}).
\newblock \bibinfo{title}{Adam: A method for stochastic optimization}.
\newblock {\it \bibinfo{journal}{arXiv preprint arXiv:1412.6980}\/}, .
\bibitem[{Kuriyan(2018)}]{kuriyan2018body}
\bibinfo{author}{Kuriyan, R.} (\bibinfo{year}{2018}).
\newblock \bibinfo{title}{Body composition techniques}.
\newblock {\it \bibinfo{journal}{The Indian journal of medical research}\/},
  {\it \bibinfo{volume}{148}\/}, \bibinfo{pages}{648}.
\bibitem[{Loshchilov \& Hutter(2016)}]{loshchilov2016sgdr}
\bibinfo{author}{Loshchilov, I.}, \& \bibinfo{author}{Hutter, F.}
  (\bibinfo{year}{2016}).
\newblock \bibinfo{title}{Sgdr: Stochastic gradient descent with warm
  restarts}.
\newblock {\it \bibinfo{journal}{arXiv preprint arXiv:1608.03983}\/}, .
\bibitem[{Majmudar et~al.(2022)Majmudar, Chandra, Yakkala, Kennedy, Agrawal,
  Sippel, Ramu, Chaudhri, Smith, Criminisi et~al.}]{majmudar2022smartphone}
\bibinfo{author}{Majmudar, M.~D.}, \bibinfo{author}{Chandra, S.},
  \bibinfo{author}{Yakkala, K.}, \bibinfo{author}{Kennedy, S.},
  \bibinfo{author}{Agrawal, A.}, \bibinfo{author}{Sippel, M.},
  \bibinfo{author}{Ramu, P.}, \bibinfo{author}{Chaudhri, A.},
  \bibinfo{author}{Smith, B.}, \bibinfo{author}{Criminisi, A.} et~al.
  (\bibinfo{year}{2022}).
\newblock \bibinfo{title}{Smartphone camera based assessment of adiposity: a
  validation study}.
\newblock {\it \bibinfo{journal}{NPJ Digital Medicine}\/},  {\it
  \bibinfo{volume}{5}\/}, \bibinfo{pages}{79}.
\bibitem[{Redmon \& Farhadi(2018)}]{YOLOv3}
\bibinfo{author}{Redmon, J.}, \& \bibinfo{author}{Farhadi, A.}
  (\bibinfo{year}{2018}).
\newblock \bibinfo{title}{Yolov3: An incremental improvement}.
\newblock \URLprefix \url{https://arxiv.org/abs/1804.02767}.
  \DOIprefix\doi{10.48550/ARXIV.1804.02767}.
\bibitem[{Shao(2014)}]{shao2014body}
\bibinfo{author}{Shao, Y.~E.} (\bibinfo{year}{2014}).
\newblock \bibinfo{title}{Body fat percentage prediction using intelligent
  hybrid approaches}.
\newblock {\it \bibinfo{journal}{The Scientific World Journal}\/},  {\it
  \bibinfo{volume}{2014}\/}.
\bibitem[{Siri(1956)}]{siri1956gross}
\bibinfo{author}{Siri, W.~E.} (\bibinfo{year}{1956}).
\newblock \bibinfo{title}{The gross composition of the body}.
\newblock In {\it \bibinfo{booktitle}{Advances in biological and medical
  physics}\/} (pp. \bibinfo{pages}{239--280}).
\newblock \bibinfo{publisher}{Elsevier} volume~\bibinfo{volume}{4}.
\bibitem[{U{\c{c}}ar et~al.(2021)U{\c{c}}ar, Ucar, K{\"o}ksal \&
  Daldal}]{uccar2021estimation}
\bibinfo{author}{U{\c{c}}ar, M.~K.}, \bibinfo{author}{Ucar, Z.},
  \bibinfo{author}{K{\"o}ksal, F.}, \& \bibinfo{author}{Daldal, N.}
  (\bibinfo{year}{2021}).
\newblock \bibinfo{title}{Estimation of body fat percentage using hybrid
  machine learning algorithms}.
\newblock {\it \bibinfo{journal}{Measurement}\/},  {\it
  \bibinfo{volume}{167}\/}, \bibinfo{pages}{108173}.
\bibitem[{Wang et~al.(1992)Wang, Pierson~Jr \& Heymsfield}]{wang1992five}
\bibinfo{author}{Wang, Z.-M.}, \bibinfo{author}{Pierson~Jr, R.~N.}, \&
  \bibinfo{author}{Heymsfield, S.~B.} (\bibinfo{year}{1992}).
\newblock \bibinfo{title}{The five-level model: a new approach to organizing
  body-composition research}.
\newblock {\it \bibinfo{journal}{The American journal of clinical
  nutrition}\/},  {\it \bibinfo{volume}{56}\/}, \bibinfo{pages}{19--28}.
\bibitem[{{World Health Organization}(2021)}]{WHO2021}
\bibinfo{author}{{World Health Organization}} (\bibinfo{year}{2021}).
\newblock {\it \bibinfo{title}{Obesity and Overweight}\/}.
\newblock \bibinfo{type}{Technical Report} World Health Organization.
\newblock \URLprefix
  \url{https://www.who.int/news-room/fact-sheets/detail/obesity-and-overweight}.

\end{thebibliography}

\end{document}